\documentclass[lettersize,journal]{IEEEtran}
\usepackage{amsmath,amsfonts}
\usepackage{algorithmic}
\usepackage{algorithm}
\usepackage{array}
\usepackage[caption=false,font=normalsize,labelfont=sf,textfont=sf]{subfig}
\usepackage{textcomp}
\usepackage{stfloats}
\usepackage{url}
\usepackage{verbatim}
\usepackage{graphicx}
\usepackage{cite}
\usepackage{adjustbox}
\usepackage{makecell}
\usepackage{multirow}
\usepackage{color}
\usepackage{array}
\usepackage{authblk}

\begin{document}

\title{PGT-Net: Progressive Guided Multi-task Neural Network for Small-area Wet Fingerprint Denoising and Recognition}


\author[1]{Yu-Ting Li}
\author[1]{Ching-Te Chiu,~\IEEEmembership{Senior Member,~IEEE,}}
\author[2]{An-Ting Hsieh} \author[3]{Mao-Hsiu Hsu} 
\author[4]{Long Wenyong} \author[1]{Jui-Min Hsu}

\affil[1]{Department of Computer Science, National Tsing Hua University}
\affil[2]{Institute of Information Systems and Applications, National Tsing Hua University}
\affil[3]{Department of Electro-Optical Engineering, National Formosa University.}
\affil[4]{FocalTech Systems Co., Ltd.}



\maketitle

\begin{abstract}

Fingerprint recognition on mobile devices is an important method for identity verification. However, real fingerprints usually contain sweat and moisture which leads to poor recognition performance. In addition, for rolling out slimmer and thinner phones, technology companies reduce the size of recognition sensors by embedding them with the power button. Therefore, the limited size of fingerprint data also increases the difficulty of recognition. Denoising the small-area wet fingerprint images to clean ones becomes crucial to improve recognition performance. In this paper, we propose an end-to-end trainable progressive guided multi-task neural network (PGT-Net). The PGT-Net includes a shared stage and specific multi-task stages, enabling the network to train binary and non-binary fingerprints sequentially. The binary information is regarded as guidance for output enhancement which is enriched with the ridge and valley details. Moreover, a novel residual scaling mechanism is introduced to stabilize the training process. Experiment results on the FW9395 and FT-lightnoised dataset provided by FocalTech shows that PGT-Net has promising performance on the wet-fingerprint denoising and significantly improves the fingerprint recognition rate (FRR). On the FT-lightnoised dataset, the false recognition rate (FRR) of fingerprint recognition can be declined from 17.75\% to 4.47\%. On the FW9395 dataset, the FRR of fingerprint recognition can be declined from 9.45\% to 1.09\%. 
\end{abstract}


\section{Introduction}
\begin{figure}[!t]
\centering
\subfloat[]{\includegraphics[height=4cm]{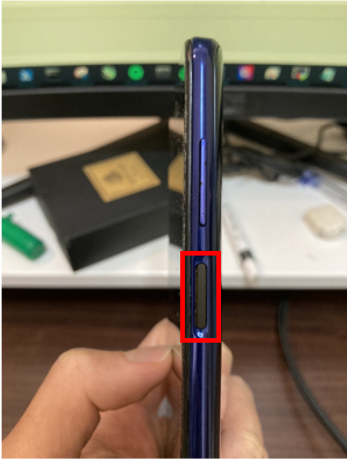}%
\label{fig:side_sensor_and_image-a}}
\hfil
\subfloat[]{\includegraphics[height=4cm]{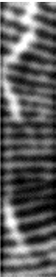}%
\label{fig:side_sensor_and_image-b}}
\caption{(a) Sensor on the side of a mobile phone that collects tiny fingerprints. (b) A fingerprint example collected by the sensor.}
\label{fig:side_sensor_and_image}
\end{figure}

\IEEEPARstart{B}{iometrics-based} security, such as fingerprint authentication, is proven to be both more secure and convenient than passwords and is widely used in our daily life. The uniqueness of fingerprints allows applications,
including electronic payment authentication, background checks, mass disaster identification, criminal identity verification, and mobile phone unlocking.
However, fingerprint images are degraded due to the noise caused by sweat, grease, or water.
Nowadays, mobile phones become thinner and the screen becomes larger, and the space for the fingerprint sensor has been compressed. Fig.~\ref{fig:side_sensor_and_image} shows an example of the sensor on the side of a mobile phone that collects tiny fingerprints of size 176 x 36. 
The size of the captured fingerprints' area is small which increases the difficulty of fingerprint recovery due to the limited information available from the input. If the fingerprint images are both small and blurry (particularly in wet conditions), the task of recovering clear fingerprints becomes more challenging.

Recently, deep learning neural networks have achieved great success in image processing. Works have been studied to apply deep learning network models to fingerprint denoising tasks. Some studies focus on latent fingerprints~\cite{Latent_Fingerprint_Enhancement_Based_on_DenseUNet, Deep_convolutional_neural_network_for_latent_fingerprint_enhancement}, which are accidentally captured at the crime scene.  In~\cite{FENET}, denoising is applied first then followed by Fingerprint pore matching. Some studies focus on fingerprints with synthetic background noises, such as~\cite{FPDMNET}. Unfortunately, to our best knowledge, few studies focus on daily applications to our lives–wet fingerprints denoising, especially tiny fingerprints with real noises caused by water, sweat, or grease. This situation may be caused by related datasets that are not easy to obtain since fingerprints are private information. Also, constructing such datasets that contain clean \& noisy fingerprint pairs requires a lot of time and effort. Also, due to the security issue, fingerprint recognition needs a very high quality of denoising performance. Recognition algorithms need to have a meager false rejection rate (FRR) to ensure that the identity can be recognized successfully. Consequently, even small changes in the fingerprint image may lead to failure in fingerprint identification.

In this work, we developed a network, progressive guided multi-task network (PGT-Net) for wet fingerprint denoising. We also attempted several techniques to improve the model’s performance. First, in the multi-task architecture, we use binary fingerprints as a supported task (guided task) to guide the main task toward more precise results. Second, our well-designed data flow makes outstanding performance, as we can get multiple helpful information through it, such as basic feature maps or the shared features between binary and non-binary fingerprints, and use the information for fingerprint denoising. Third, the novel residual scaling can help stabilize the training loss. Fourth, the binary progressive guided multi-task can produce precise binary fingerprint contour, which allows the main task to know how to restore the fingerprint better. Last but not least, the optimized loss function adds structural similarity index measure (SSIM) loss and Laplacian loss, which significantly improves the denoising results. In this work, we present the PGT-Net-block-84 model and PGT-Net-block-edge for heavy and light weight applications.

\begin{figure}[!t]
\centering
\includegraphics[width=\linewidth]{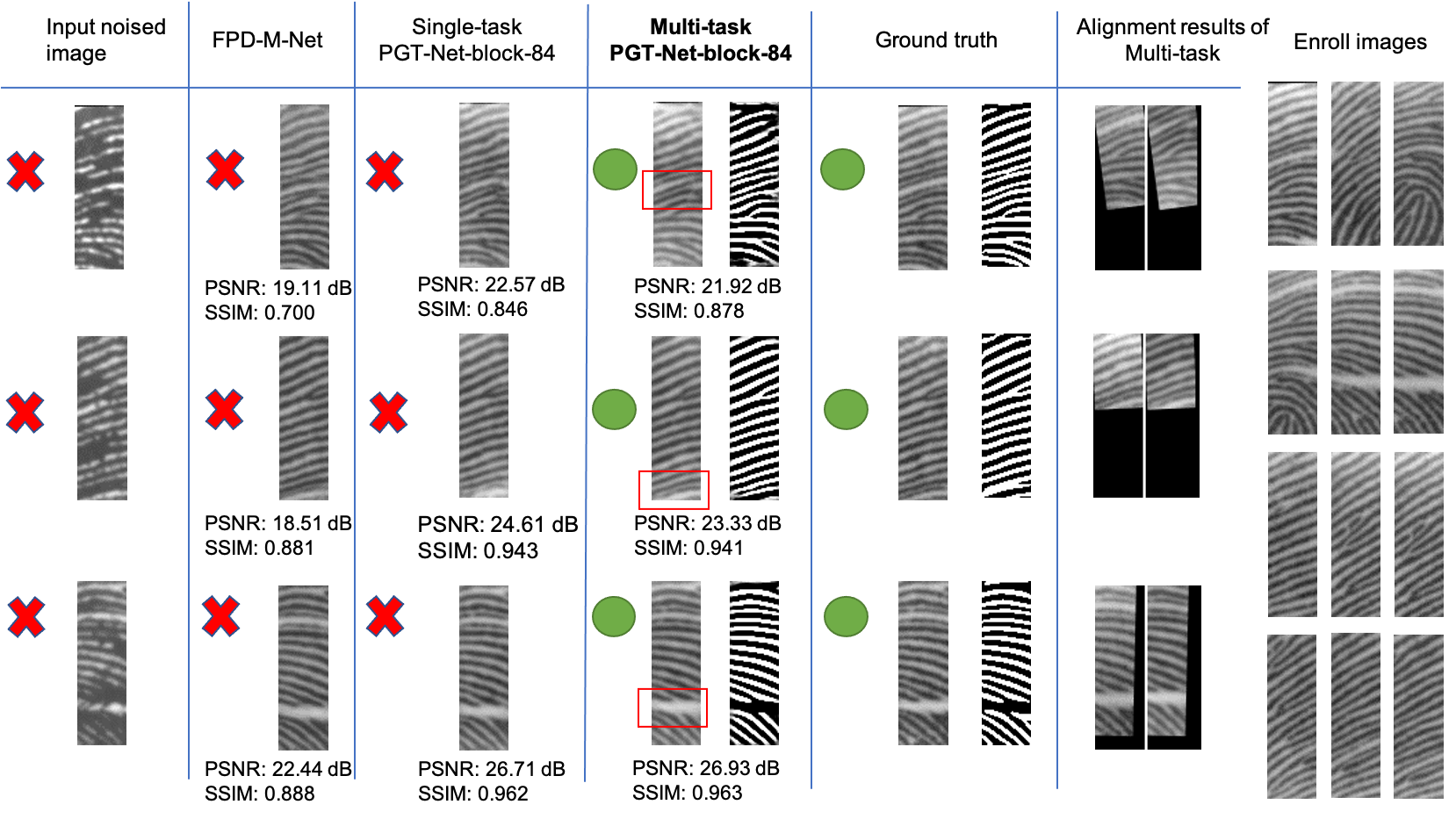}
\caption{Real-world Fingerprint denoising and recognition results with different model on FT-lightnoised dataset.}
\label{fig:multisingle_FPDM_result}
\end{figure}

Fig.~\ref{fig:multisingle_FPDM_result} shows an example of the denoised fingerprints that fail to be recognized with their corresponding input noisy image and output ground truth image. A red cross means a fingerprint fails to be identified, while a green circle represents a fingerprint that can be identified successfully. As shown in  The noisy fingerprints in the first column fail to be recognized since the finger is covered with water, grease, or sweat. The images in the second and third columns are the denoised results of the FPD-M-Net~\cite{FPDMNET} and single-task PGT-Net-block-84 model, the denoised quality is quite well that the denoised fingerprint is very similar to the ground truth in the fourth column, but it still cannot be recognized. While our proposed multi-task PGT-Net-block-84 neural network not only removes real-world noise but keeps the detail of the contour to succeed in passing the recognition, as shown in the fourth column.

\begin{figure}[!t]
\centering
\includegraphics[width=0.45\linewidth]{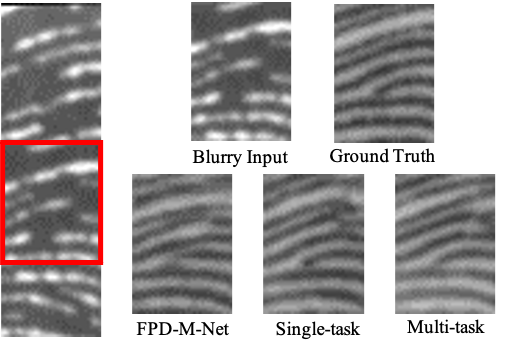}
\caption{Enlarged critical parts of the fingerprints that affect the recognition results.}
\label{fig:multisingle_FPDM_result_large}
\end{figure}

To further show the different results between models, we enlarged the critical parts of the fingerprints that affect the recognition results, as shown in Fig.~\ref{fig:multisingle_FPDM_result_large}. The proposed multi-task PGT-Net-block-84 has the capability of not only denoising the fingerprint, and also making it precise and correct.

To be summarized, the contributions of our work are listed below:

\begin{itemize}
    \item The proposed model can be used the wet fingerprints with tiny image sizes and significantly reduce the false rejection rate (FRR).
    \item The convenience of fingerprint recognition has been greatly improved.
        \begin{itemize}
            \item The FRR of the FT-lightnoised dataset (real noise) can be improved from 17.75 \% to 4.47 \%
            \item The FRR of the FW9395 dataset (synthetic noise) can be improved from 9.45\% to 1.09 \%
        \end{itemize}
  \item Outperform state-of-the-art network models on fingerprint denoising.
\end{itemize}

\section{Related Work}
\label{sec:related_works}

Fingerprints enhancements related technologies have been developed for decades, and many traditional methods or algorithms have been introduced. Such as Gabor-filtering-based fingerprint enhancements~\cite{fingerprint_enhancer, Orientation_field_estimation_for_latent_fingerprint_enhancement, A_modified_Gabor_filter_design_method_for_fingerprint_image_enhancement}, orientation field estimation methods~\cite{Orientation_field_estimation_for_latent_fingerprint_enhancement, Localized_dictionaries_based_orientation_field_estimation_for_latent_fingerprints, A_multiscale_directional_operator_and_morphological_tools_for_reconnecting_broken_ridges_in_fingerprint_images_Pattern_Recognit, Systematic_methods_for_the_computation_of_the_directional_fields_and_singular_points_of_fingerprints, Fingerprint_recognition_by_combining_global_structure_and_local_cues, FOMFE}, total variation (TV) model~\cite{A_comparison_of_three_total_variation_based_texture_extraction_models, Adaptive_Directional_Total_Variation_Model_for_Latent_Fingerprint_Segmentation}. In general, these conventional methods may not perform well when heavy noise exists, which leads to poor fingerprint recovery or recognition performance. Consequently, it’s still beneficial for deep learning training data pre-processing~\cite{Deep_convolutional_neural_network_for_latent_fingerprint_enhancement}.

As described previously, small-area wet fingerprint denoising is difficult since there is limited input information, and it needs high precision so that the denoised fingerprints can be recognized. 
Most of current researches either have a poor denoising performance, or the denoised fingerprint contour is not precise enough for real wet fingerprint images. Below are the description of deep learning with different neural network models for fingerprint denoising.

\subsection{Residual Neural Networks} 
Residual learning and residual neural networks (ResNet) were invented by He et al.~\cite{he2016deep} to solve the problem of performance degradation as a network model's depth increases. This learning mechanism adds the extracted features and the output from previously convolution layers. 
It  achieves remarkable success in the computer vision field.

The PFE-Net~\cite{FENET} makes use of the residual architecture to improve its performance. To reconstruct the input images, they make use of the residual structure to learn the local features. Also, the scaling is decreasing layer by layer between the residual blocks, which makes the PFE-Net able to learn the fingerprint features robustly.

\subsection{U-Net}
The U-Net~\cite{UNet} is an autoencoder architecture that has been used widely. It was proposed for the semantic segmentation task. The architecture has two main components, an encoder path, and a symmetric decoder path. The encoder is used to capture the context, and the decoder is used to estimate the segmentation. The symmetric architecture concatenates feature maps at the same level, thus improving the feature map's reusability and reducing the information loss during the encoding or decoding process.

FusionNet~\cite{FusionNet} is a U-Net-based architecture, which focuses on extracting cellular membrane segmentation in electron microscopy (EM) images, and had achieved great success in the EM segmentation tasks. Jung Yoon Bae et al.~\cite{FINGERPRINT_IMAGE_DENOISING_AND_INPAINTING_USING_CONVOLUTIONAL_NEURAL_NETWORK} proposed a network model for fingerprint denoising that is based on FusionNet. DenseUNet~\cite{Latent_Fingerprint_Enhancement_Based_on_DenseUNet} is also based on U-Net and focuses on latent fingerprint enhancement. Also, it uses dense blocks to improve the information flow between layers.

\subsection{M-Net}
The M-Net~\cite{MNet} is a U-net based model which is modified for better segmentation. The main difference is that two side paths are added to the two main encoding and decoding paths. The model preserves the input details with these two legs in case the downsampling operator drops image details. Besides, skip connections between corresponding encoding and decoding layers provide sufficient information and better features. 

The FPD-M-Net~\cite{FPDMNET} is a model that focuses on latent fingerprint denoising. It is modified from M-Net to fit the fingerprint denoising 
and inpainting tasks. The segmentation-based architecture demonstrated the capability to handle both denoising and inpainting tasks for fingerprint images simultaneously. The proposed model outperformed both the U-net and the baseline model provided in the competition. The model exhibited robustness to challenges such as strong background clutter and weak signals and effectively performed automatic filling. The findings also emphasized the importance of sensor-specific training for improved results when dealing with images acquired from diverse sensors.

\subsection{Multi-tasks Learning}
Multi-tasks learning is a technique that trains multiple tasks together to improve the accuracy or reduce the parameters. Such a technique has been applied to different domains in deep learning~\cite{Deep_convolutional_neural_network_for_latent_fingerprint_enhancement, An_overview_of_multi_task_learning_in_deep_neural_networks, Multi_task_learning_using_uncertainty_to_weigh_losses_for_scene_geometry_and_semantics, Regularized_multi_task_learning, Recurrent_neural_network_for_text_classification_with_multi_task_learning, Hyperface, Adversarial_multi_task_learning_for_text_classification, Facial_landmark_detection_by_deep_multi_task_learning}, such as text classification~\cite{Multi_task_learning_using_uncertainty_to_weigh_losses_for_scene_geometry_and_semantics, Recurrent_neural_network_for_text_classification_with_multi_task_learning}, or image processing~\cite{Deep_convolutional_neural_network_for_latent_fingerprint_enhancement, An_overview_of_multi_task_learning_in_deep_neural_networks, Multi_task_learning_using_uncertainty_to_weigh_losses_for_scene_geometry_and_semantics, Regularized_multi_task_learning, Hyperface, Facial_landmark_detection_by_deep_multi_task_learning}.

Conventionally, deep learning models reach their target directly. For example, for fingerprint denoising, most of the studies get the denoised output fingerprint directly. While in our work, the denoised binary output fingerprints will be obtained first, and then the denoised non-binary output fingerprints, which are our main target.

FingerNet~\cite{Deep_convolutional_neural_network_for_latent_fingerprint_enhancement} introduced a multi-task deep learning fingerprint denoising network model. Some significant differences exist between our works and FingerNet~\cite{Deep_convolutional_neural_network_for_latent_fingerprint_enhancement}. 

First of all, although both our work and ~\cite{Deep_convolutional_neural_network_for_latent_fingerprint_enhancement} are multi-task, the weights of the tasks are different in our work. In ~\cite{Deep_convolutional_neural_network_for_latent_fingerprint_enhancement}, two tasks exist in the model, one is the orientation task, and another is the enhancement task; these two tasks are equally important and produce equally important outputs. While in our work, we also have two tasks, one task is the “progressive guided multi-task,” which has a lighter weight during training, and the other one is the “main task,” which has a heavier weight during training.  We  focus more on the output of the main task, the output of the supported task is only for guiding the main task to achieve better denoising performance.

Second, the output of the supported task in the PGT-Net has a concatenate path to the input of the main task branch. Such concatenation further helps the main task improve its denoising performance since the denoised binary fingerprint can provide some helpful information such as fingerprints orientation or the contour of ridges and valleys in the fingerprints. While in ~\cite{Deep_convolutional_neural_network_for_latent_fingerprint_enhancement}, the task output wasn’t utilized for further improvements.

\section{Network Model}
\label{ch:method} 

The PGT-Net is a multi-task network model. It takes noisy non-binary inputs and then separates them into two different paths to produce two outputs: denoised binary outputs and denoised non-binary contour outputs. Fig.~\ref{fig:PGT_dataflow} shows the data flow and the critical ideas of our proposed method. Besides, Fig.~\ref{fig:PGT-Net-block-84_dataflow} displays the complete architecture of our proposed model. There are four main data flow blocks in the PGT-Net. 

\subsubsection{Produce the Basic Feature Map (BFM)}
The basic feature map is the important feature map that contains extracted features of the input image. It can also provide information to reduce feature loss during the following denoising process. BFM will be concatenated to the output of data flow block C and the output of data flow block D.
  
\subsubsection{Processing Shared Features}
There are some shared features between the non-binary fingerprints and binary fingerprint contour. For example, denoised binary and non-binary fingerprints must have clear ridges and valleys to achieve a higher identification rate after denoising. Consequently, this step, “Processing shared features”, is necessary.

\subsubsection{Processing Progressive Guided Features}
As described previously, the denoised binary and non-binary fingerprints must have a precise contour to have a higher identification rate. The denoised binary fingerprint image contains accurate information about the ridges and valleys of the fingerprints, which is very helpful for the non-binary fingerprint image denoising.

\subsubsection{Processing Main Task’s Features}
 At this step, the data flow block D receives the output of data flow block B,and the denoised binary output as its input. Those feature maps contain valuable information that can be used to denoise the non-binary fingerprints, which is same with our main target.
 
\begin{figure}[!t]
\centering
\includegraphics[width=\linewidth]{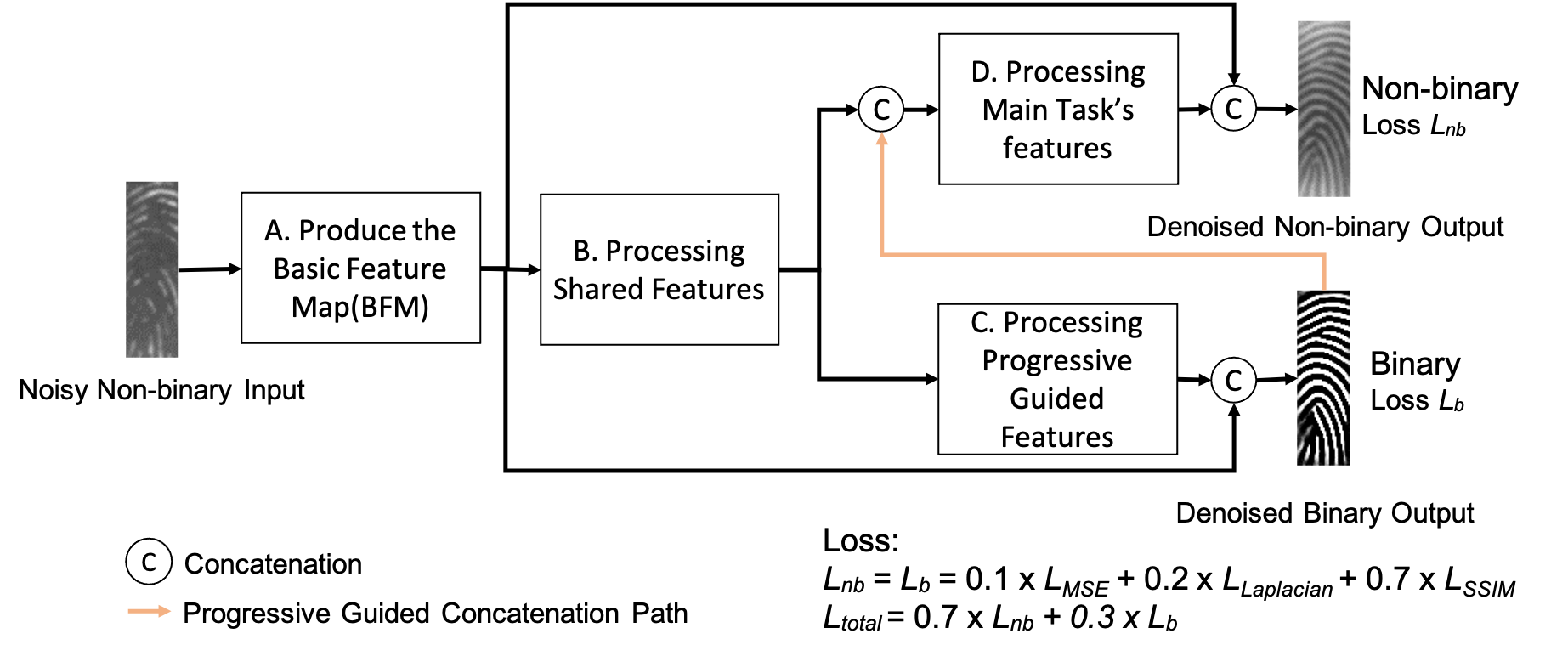}
\caption{The data flow of the proposed PGT-Net model.}
\label{fig:PGT_dataflow}
\end{figure}

 Our dataflow uses concatenation operations, as shown in Fig.~\ref{fig:PGT_dataflow}. The concatenate operations are used to increase the number of feature maps. And the concatenate operation can be described precisely and mathematically with the concept of “set”. 
 
\begin{figure}[h]
    \centering
    \includegraphics[width=\linewidth]{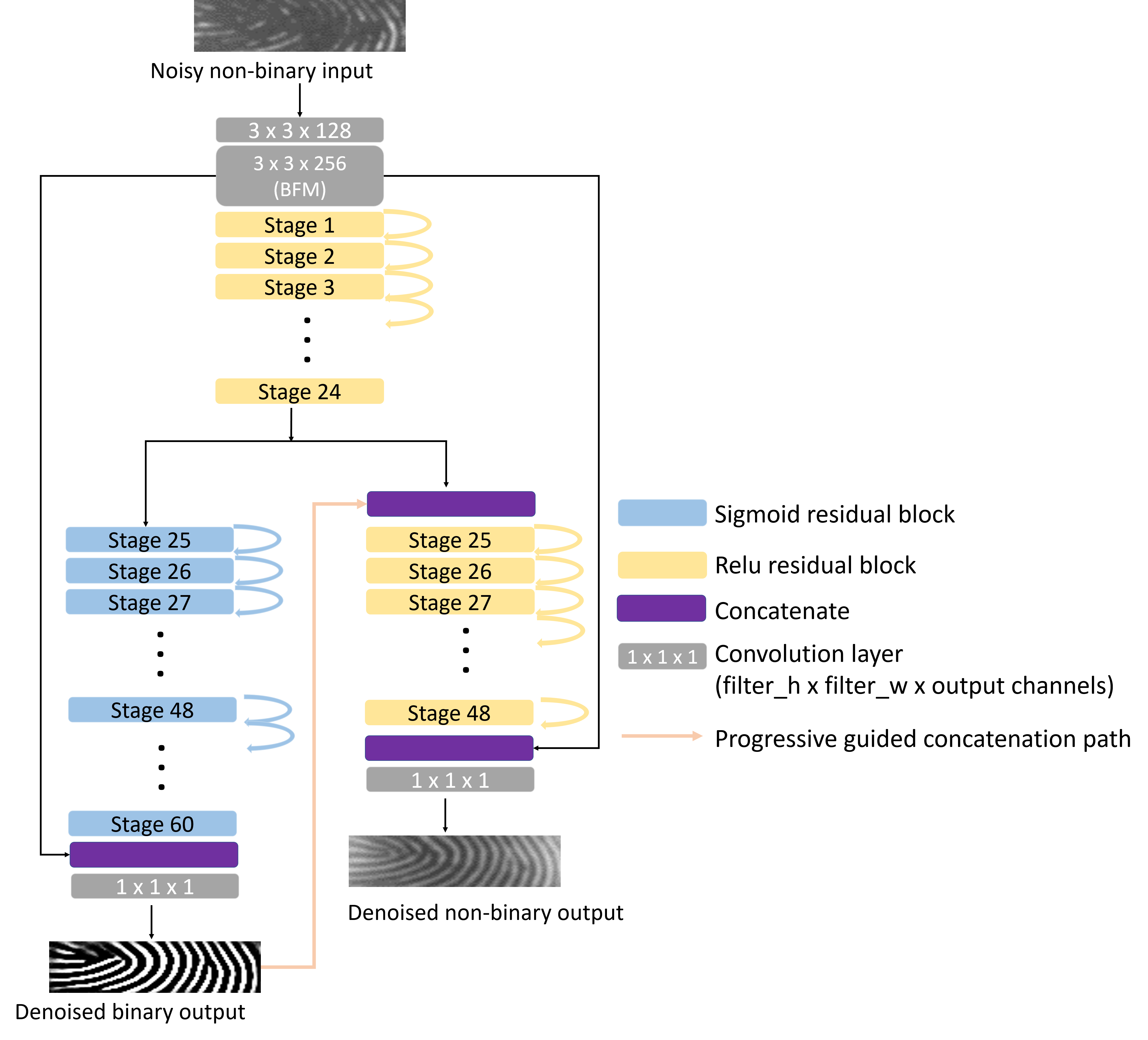}
    \caption{The network architecture of multi-task PGT-Net-block-84.}
    \label{fig:PGT-Net-block-84_dataflow}
\end{figure}

\subsection{PGT-Net-block-84}
\label{sec:PGT_Net_block_84}
To further improve the performance of fingerprint recognition after denoising, PGT-Net-block-84 is proposed. Fig.~\ref{fig:PGT-Net-block-84_dataflow} shows the detailed architecture of the proposed PGT-Net-block-84.

\begin{figure}[!t]
\centering
\includegraphics[scale=0.45]{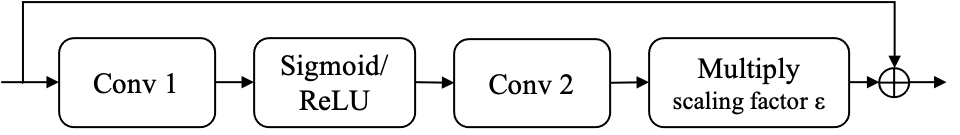}
\caption{The architecture of a residual block.}
\label{fig:residualBlock}
\end{figure}

\begin{itemize}
    \item The first two convolution layers are used to produce the BFM.
    \item Residual blocks stage 1-24 are used to the shared features, and Fig.~\ref{fig:residualBlock} shows the architecture of a residual block. Further information about residual blocks will be discussed in section~\ref{Residual_blocks}.
    \item Residual blocks stage 25-60 of the binary branch  are used to process the
progressive guided features.
    \item  Residual block stages 25-48 of the non-binary branch are used to process the
main task's features.
    \item PGT-Net-block-84 has 84 residual blocks in total.
\end{itemize}

And detailed information about the architecture will be discussed in sections~\ref{Residual_blocks} to ~\ref{Optimized_loss_function}.

\subsubsection{Residual Blocks}
\label{Residual_blocks}
The architecture of residual blocks is shown in Fig.~\ref{fig:residualBlock}. Each residual block contains two convolution layers, and the choice between 2 activation functions, Sigmoid and Relu, is based on the residual block's task. For those residual blocks that handle the non-binary information, Relu will be chosen. While for those residual blocks that handle binary information, Sigmoid will be selected, as it could help the pixel value converge to either one or zero. For example, In PGT-Net, a binary task guides the non-binary task to produce better results. After processing through the shared features layers, the left side of the fork handles the binary task, and the right side deals with the non-binary task. 

Also, there is a scaling factor \(\varepsilon\) in each residual block, \(\varepsilon\) is set to \(0.01 \times \alpha \) initially, and it will decay 0.01 as the stage increase by one, as shown in Eq.~\ref{eq:eplison}. The residual scaling is proven helpful when training a deeper network, and such a technique makes the model more robust~\cite{Enhanced_deep_residual_networks_for_single_image_super_resolution, Deep_residual_learning_for_image_recognition, Stable_Pore_Detection_for_High_Resolution_Fingerprint_based_on_a_CNN_Detector}.

\begin{figure}[h]
    \centering
    \includegraphics[height=7cm]{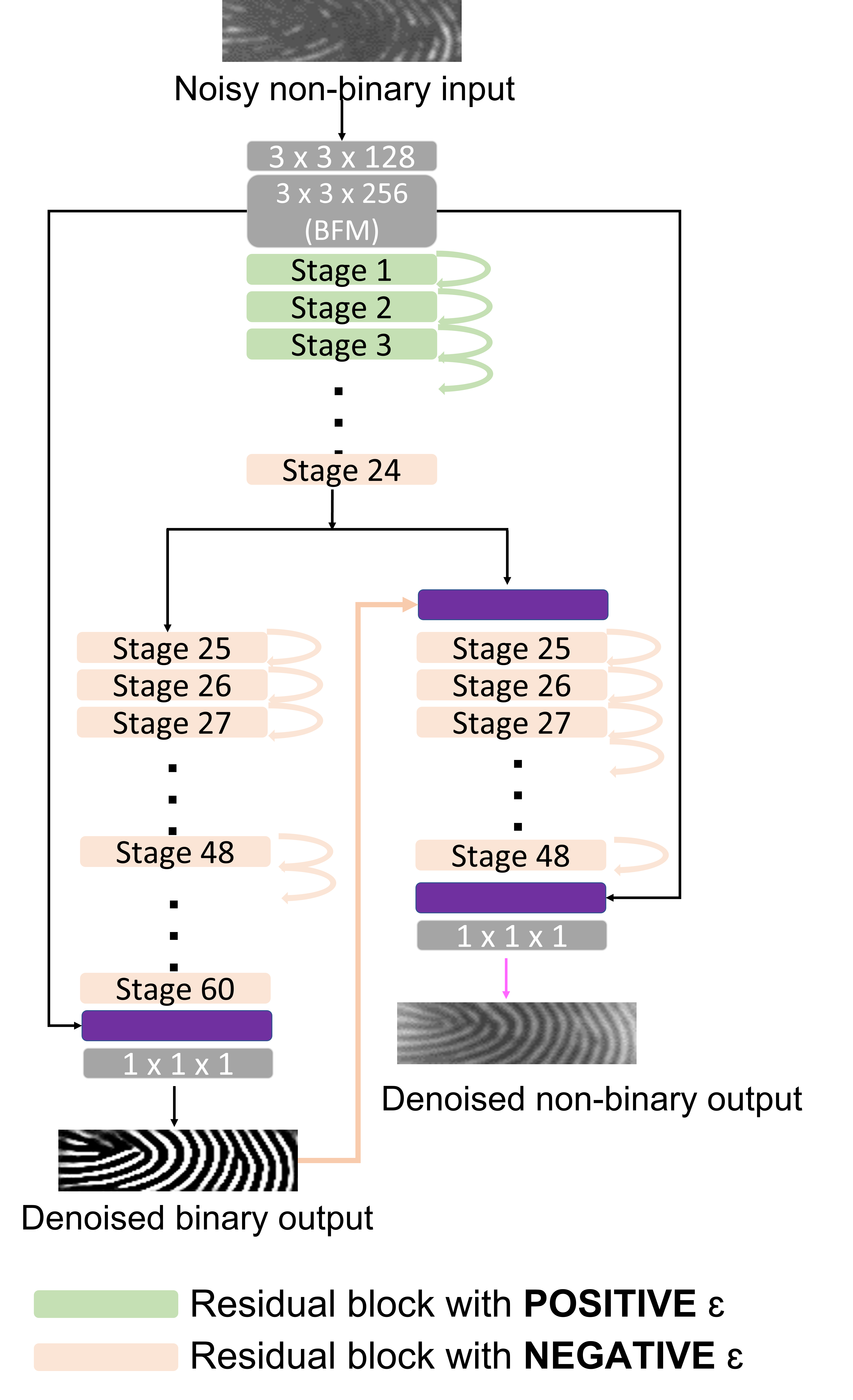}
    \caption{Residual blocks with positive or negative scaling factor\(\varepsilon\).}
    \label{fig:PGT-Net-block-84_scale}
\end{figure}

Note that the residual scaling factor \(\varepsilon\) could be either positive or negative. For example, the residual block at stage 15 will have \(\varepsilon\) = 0.01 \(\times\) (24 - 15) = 0.09, the residual block at stage 30 will have \(\varepsilon\) = 0.01 \(\times\) (24 - 30) = -0.06. We use positive scaling to extract fingerprint features and use negative scaling for denoising. Take PGT-Net-block-84 as an example, stage 1 to 23 are responsible for extracting shared features, so its scaling factor \(\varepsilon\) is set to be positive. Stage 24 to 60 are responsible for the binary or non-binary fingerprint image denoising, so they have a negative scaling factor \(\varepsilon\). Fig.~\ref{fig:PGT-Net-block-84_scale} shows the residual blocks with positive scaling factor \(\varepsilon\) or negative scaling factor \(\varepsilon\). To avoid the scaling factor being zero, the residual scaling after stage 24 is subtracted 0.01 explicitly. Experiments have shown that such a residual scaling technique can achieve better performance, further analytic data can be referred to in section ~\ref{sec:Residual_scaling_factor}.

\begin{equation}
\label{eq:eplison} 
\varepsilon= 
\begin{cases}
    0.01(\alpha - \text{current stage}),& if \text{current stage} < \text{24} \\
    0.01 (\alpha - \text{current stage}) - 0.01 , & otherwise
\end{cases}
\end{equation}

\subsubsection{Binary Progressive Guided Task}
We chose binary fingerprints as our guided task because the textures in binary fingerprints are more precise. For example, the textures are more evident in binary fingerprints than in non-binary ones. Thus binary fingerprints can provide a better quality of fingerprint contour information, which can guide the main task to find the critical information, such as the ridges and valleys of fingerprints, thus reaching a better denoising performance.

\subsubsection{Optimized Loss Function}
\label{Optimized_loss_function}
Laplacian Operator is a differential operator widely used in image edge detection. We add Laplacian loss to our loss function because we need to emphasize the edge information between edges and valleys of a fingerprint so that the model learns how to restore the orientation of a noisy fingerprint.

Eq.~\ref{eq:lap_operator} shows the definition of the Laplacian operator, the Laplacian of f is the sum of all unmixed second partial derivatives in the Cartesian coordinates \(x_i\).

\begin{equation}
\label{eq:lap_operator}
\Delta f=\sum_{i=1}^{n}\frac{\partial^2 f}{\partial x_i^2}
\end{equation}

In practice, we usually use the Laplacian filter to convolve with the image to get the gradient of the image. Eq.~\ref{eq:lap_discrete} shows the discrete form of the Laplacian operator, namely, the Laplacian filter~\cite{FENET}.

\begin{equation}
\label{eq:lap_discrete} 
Laplacian\;filter=
\begin{bmatrix}
-1 & -1 & -1\\
-1 & 8 & -1\\
-1 & -1 & -1\\
\end{bmatrix}
\end{equation}

\begin{figure}[!t]
\centering
\subfloat[]{\includegraphics[height=2.3cm]{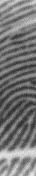}%
\label{fig:lap_filter_example-a}}
\hfil
\subfloat[]{\includegraphics[height=2.3cm]{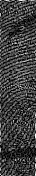}%
\label{fig:lap_filter_example-b}}
\hfil
\subfloat[]{\includegraphics[height=2.3cm]{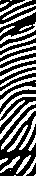}%
\label{fig:lap_filter_example-c}}
\hfil
\subfloat[]{\includegraphics[height=2.3cm]{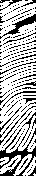}%
\label{fig:lap_filter_example-d}}
\caption{Fingerprint convolves with the Laplacian filter. (a) normal fingerprint (b) normal fingerprint with Laplacian filter (c) binary fingerprint (d) binary fingerprint with Laplacian filter}
\label{fig:lap_filter_example}
\end{figure}

Fig.~\ref{fig:lap_filter_example-a}, b shows the result of a non-binary fingerprint convolves with the Laplacian filter. Fig.~\ref{fig:lap_filter_example-c}, d shows the result of a binary fingerprint convolves with the Laplacian filter. As you can see, compared to the non-binary result, binary fingerprints emphasize the gradient, which makes the model easier to learn how to restore the texture of fingerprints.

Eq.~\ref{eq:L_mse},~\ref{eq:L_lap},~\ref{eq:ssim},~\ref{eq:L_ssim},~\ref{eq:single} shows the loss function of a single-task used in PGT-Net. In addition to the mean squared error (MSE), we also add SSIM \& Laplacian loss. SSIM in loss function can help to improve the performance in terms of SSIM in our experiments. And Laplacian loss can calculate the gradient of fingerprints, which allows the model to identify the orientations of fingerprints~\cite{FENET}. Eq.~\ref{eq:total} shows the weighting of the loss function are empirically set as 0.7 for non-binary task (main task) and 0.3 for binary task (guided/supported task).

\begin{equation}
\label{eq:L_mse} 
{L_{MSE}} = \frac{1}{N}\sum_{ij}(x_{ij} - y_{ij})^2
\end{equation}

\begin{equation}
\label{eq:L_lap} 
{L_{Laplacian}} = \sum_{ij}(Lap\;filter(x_{ij})- Lap\;filter(y_{ij}))^2
\end{equation}

\begin{equation}
\label{eq:ssim} 
{SSIM(x,y)} = \frac{(2\mu_{x}\mu_{y} + C_1)(2\sigma_{xy}\mu_{y} + C_2)}{(\mu_{x}^2 + \mu_{y}^2 + C_1)(\sigma_{x}^2 + \sigma_{y}^2 + C_2)}
\end{equation}

\begin{equation}
\label{eq:L_ssim} 
{L_{SSIM}} = 1 - SSIM(x, y)
\end{equation}

\begin{equation}
\label{eq:single} 
Single\;task\;loss = 0.1{L_{MSE}}+0.2{L_{Lap}}+0.7{L_{SSIM}}
\end{equation}

\begin{equation}
\label{eq:total} 
\begin{split}
Total\;Loss =  0.3 \times{binary\;task\;loss}\\ 
+ 0.7\times {non\_binary\;task\;loss}
\end{split}
\end{equation}

\section{Network Model on Edge Devices}
\label{ch:PGT_Net_on_edge_devices} 
The fingerprint-related applications are usually on edge devices with limited computing resources. As a result, the denoising algorithm can not be very complicated or needs heavy computations.

PGT-Net is also a neural network model with high scalability. One can adjust the number of the residual blocks or the architecture easily. As long as one follows the data flow described in section~\ref{ch:method}, they can also get a model with excellent performance with acceptable downgrade but lower complexity.

PGT-Net-Edge provides an example of simplifying the model to reduce the computation resources on edge devices. One can follow similar steps to streamline their model according to their demands. Fig.~\ref{fig:PGT_Net_Edge_and_PGT_single_task}(a) shows the architecture of the PGT-Net-Edge model. The PGT-Net-Edge reduces the complexity be three ways described below.

\subsubsection{Residual Blocks Quantity Reduction}
The number of residual blocks can be easily reduced, and the reduction can depend on the application. For example, in PGT-Net-Edge, the number of sigmoid residual blocks is reduced to zero, which handles the binary supported task initially.

\subsubsection{Residual Blocks Output Channel Reduction}
The output channels of residual blocks can also be adjusted according to the application. And reduce the output channel can significantly reduce the number of parameters. For example, in the original PGT-Net, all residual blocks have 64 output channels. While in PGT-Net-Edge, stage 1 to 28 has 32 output channels, and stage 29 to 32 has only 16 output channels. 

\subsubsection{Dynamic Fixed Point Quantization}~\cite{Real_Time_Block_Based_Embedded_CNN_for_Gesture_Classification_on_an_FPGA} has proposed an algorithm that quantifies the floating-point weights or feature maps to dynamic-fixed-point. Compared to static-fixed-point quantization, dynamic-fixed-point quantization is more precise under the same precision and thus performs better. The experimental result of PGT-Net-Edge after quantization will be shown in section~\ref{sec:Performance_summarization}.

\section{Datasets}
\label{sec:Datasets}

\subsection{Denoising Datasets}
\label{sec:Denoising_datasets}
We use the FW9395 and FT-lightnoised two datasets for training. Both of the datasets are divided into training, validation, and testing and provided by FocalTech. It is described in Table~\ref{tab:dataset}. Both of the datasets consist of a pair of blurry wet and ground-truth fingerprint images.

\subsubsection{FT-lightnoised Dataset}
The FT-lightnoised dataset is collected by the Focaltech optical image sensor. The sensor captures the entire clean and wet fingerprints. After aligning the clean and wet pair, we obtain the small-area fingerprint images by cropping the entire images to size 176 x 36. Fig.~\ref{fig:dataset_examples-a} shows an example image pair.

\subsubsection{FW9395-synthetic Dataset}
The dataset, FW9395-synthetic dataset, we used in the research is collected by the Focaltech capacitive image sensor. Because of the limitation of the hardware, the capacitive image sensor is hard to produce aligned wet \& clean fingerprint pairs. So we used Gaussian noise to simulate the wet fingerprints noise. Fig.~\ref{fig:dataset_examples-b} shows an example pair of images in the FW9395-synthetic dataset. And the synthetic process was introduced in Section~\ref{sec:synthesize}.

\subsection{Data Preprocessing}
\label{sec:Data_preprocessing}
\subsubsection{Generate Binary Fingerprint Images}
\begin{figure}[!t]
\centering
\subfloat[]{\includegraphics[height=2.3cm]{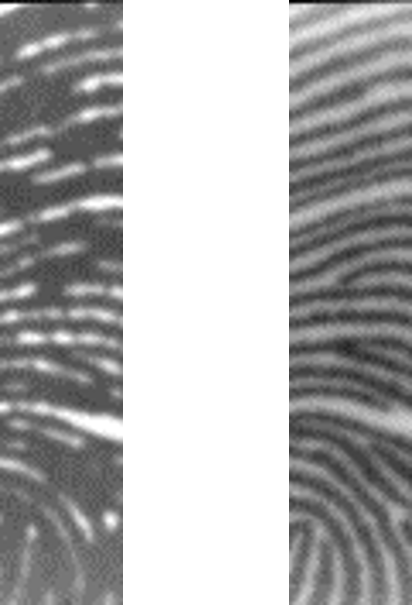}%
\label{fig:dataset_examples-a}}
\hfil
\subfloat[]{\includegraphics[height=2.3cm]{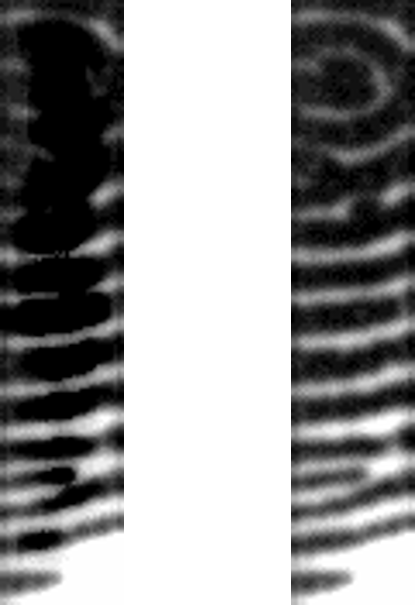}%
\label{fig:dataset_examples-b}}
\hfil
\subfloat[]{\includegraphics[height=2.3cm]{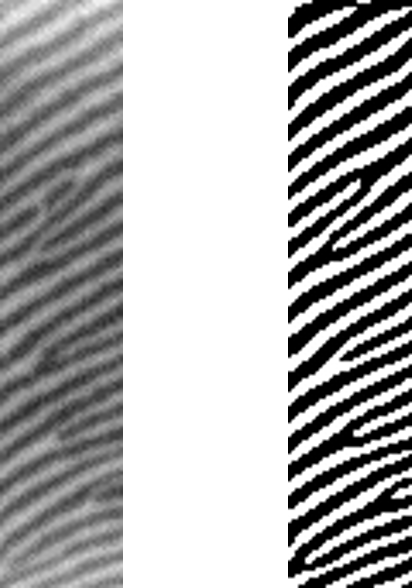}%
\label{fig:dataset_examples-c}}
\caption{Fingerprint pairs on different datasets. (a) FT-lightnoised wet and clean pair (b) FW9395 wet and clean pair (c) clean fingerprint and corresponding binary fingerprint}
\label{fig:dataset_examples}
\end{figure}

Binary fingerprints are used as supporting information to improve the wet fingerprints' denoising performance in our work. 
Friction ridges of a finger contain important features of fingerprint, therefore, recovering correct ridges plays a crucial role in the task. Thus, we generated reliable binary data from the original ground truth by fingerprint enhancer~\footnote{Fingerprint enhancer,\url{https://github.com/Utkarsh-Deshmukh/Fingerprint-Enhancement-Python}.} which was based on Hong's research~\cite{fingerprint_enhancer}, so that these binary data will lead the model to learn better ridges features. Fig.~\ref{fig:dataset_examples-c} shows an example of a non-binary fingerprint and its corresponding binary fingerprint. The ridges of the fingerprints will set to 1, while the valleys of which will be 0.

\subsubsection{Synthesize Blurry Fingerprint Images}
\label{sec:synthesize}
In FW9395, images are collected by a capacitive sensor, which has the difficulty to get aligned wet blurry and clean normal fingerprint data. To have correspondence between noised fingerprint and normal fingerprint for neural network training, we need to synthesize blurry fingerprint images from clean normal fingerprint images through the following steps.

\begin{figure}[ht]
    \centering
    \includegraphics[width=0.8\linewidth]{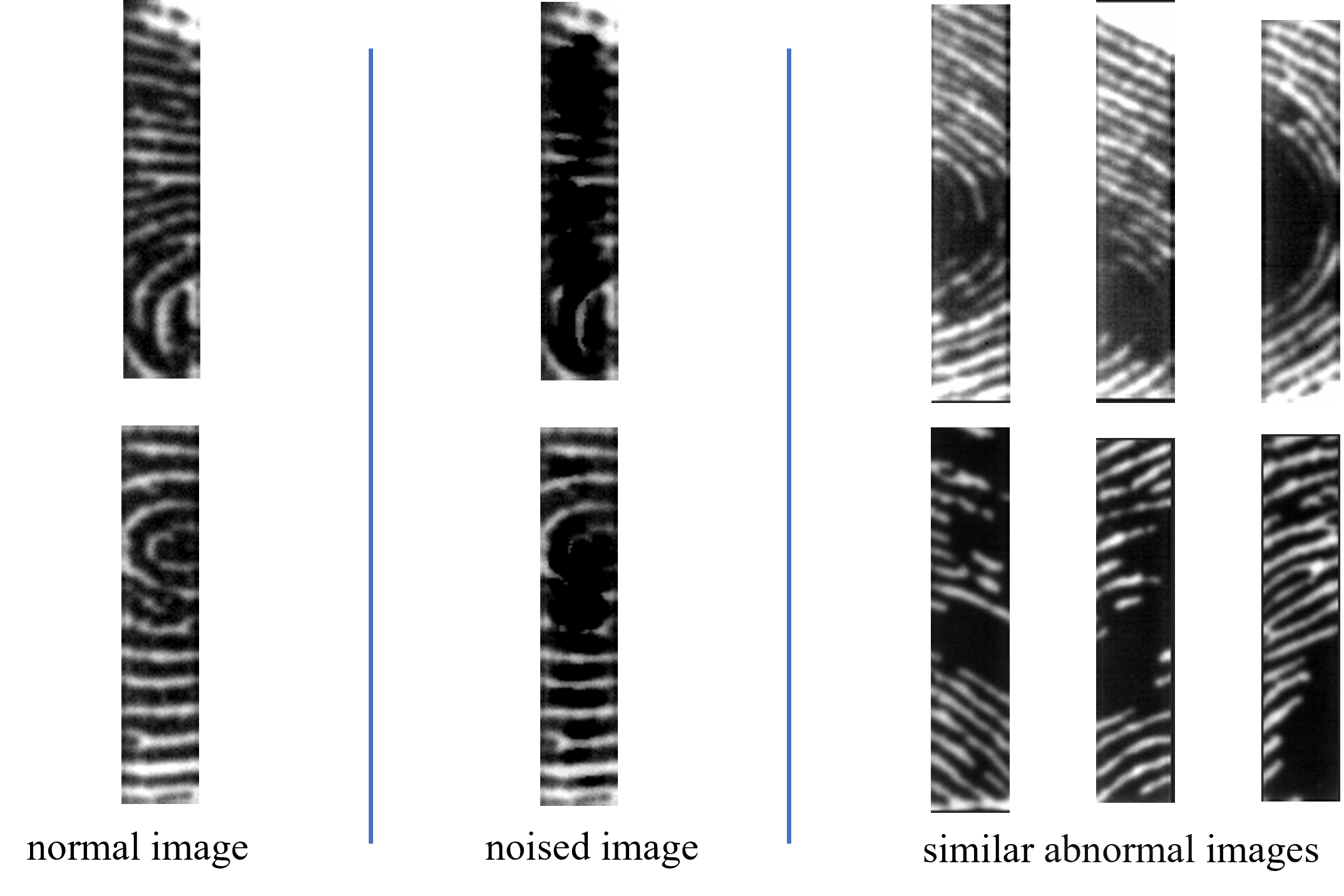}
    \caption{FW9395 synthetic data examples.}
    \label{fig:FW9395}
\end{figure}

\begin{enumerate}
   \item Binarize the fingerprint.
   \begin{enumerate}
       \item Ridges in the original image will have a pixel value of 255 in the binarized image (white)
       \item Valleys in the original image will have a pixel value of 0 in the binarized image (black)
   \end{enumerate}
    \item Create Gaussian kernel with size 13, 15, 17, 19, 21, std\_dev = 1
    \item Set the following parameter 
    \begin{enumerate}
        \item Appearance probability of noise = $0.2$
        \item Darkness of noise = $-0.2$ (the smaller the value, the darker the noise)
        \item Darkness range = $(-0.01\;,\;0.01)$
    \end{enumerate}
    \item Select a pixel on the ridge (x, y), apply Gaussian kernel.
    \item Set darkness\_value = $(darkness\;of\;noise) + random(darkness\;range)$ 
    \item Update Gaussian kernel with the multiplication of Gaussian kernel and darkness\_value 
    \item Apply Gaussian kernel to the chosen area.
 \end{enumerate}
 The synthetic results are displayed in Fig.~\ref{fig:FW9395}.
 
\renewcommand{\arraystretch}{1.8}
\begin{table}[!t]
    \centering
    \caption{Dataset comparisons\label{tab:dataset}}
    \begin{adjustbox}{width=\linewidth, center}
    \begin{tabular}{|c|c|c|c|c|}     
        \hline
        Dataset &  Scanner & \makecell{Train Images} & \makecell{Validation Images}  & \makecell{Test Images}\\
        \hline
        FT-lightnoised & Optical   &   \makecell{40,962} & 5,120  & 5,706
        \\
        \hline
        FW9395 &  Capacitive   &  \makecell{21,660} & 2,708  & 2,708  \\
        \hline
    \end{tabular}
    \end{adjustbox}
\end{table}

\subsection{Recognition Datasets}
\label{sec:Recognition_tools_and_datasets}
The fingerprint recognition tool is provided by FocalTech whose algorithm is based on the referred to SIFT~\cite{SIFT}, which is the same as the fingerprint recognition software applied to smartphones. The tool requires two inputs: the enrolled and the identified fingerprint images. The enrolled images are clean fingerprints and will be registered on each finger into the FocalTech recognition tool. The identified images are noisy fingerprints with water, grease, or sweat. After the identified fingerprint images are restored by the model, they will be put into the recognition tool to examine if the restored outputs are able to be recognized. 
The FT-lightnoised recognition dataset contains fingerprints from four people with different fingers. The FW9395 recognition dataset contains fingerprints from six people with different fingers. Table~\ref{tab:recognition_dataset} summarizes the detailed information of the datasets and their corresponding FRR when the denoising algorithm is not applied.

\begin{figure}[!t]
\centering
\includegraphics[width=\linewidth]{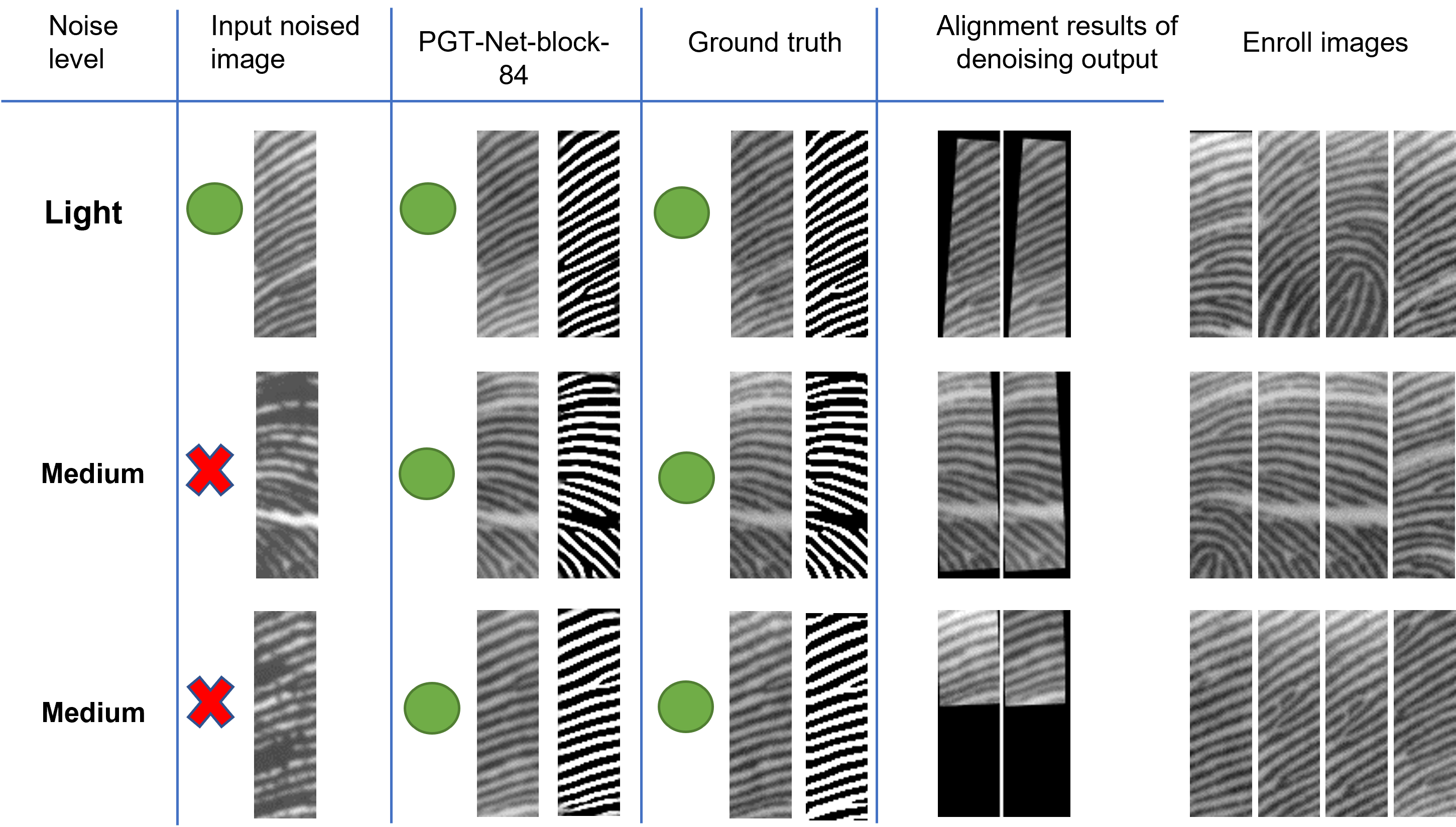}
\caption{PGT-Net-block-84 denoised and recognition
results on the FT-lightnoised dataset (light and medium noise).}
\label{fig:ft_light_summarize_img_lm}
\end{figure}

\begin{figure}[!t]
\centering
\includegraphics[width=\linewidth]{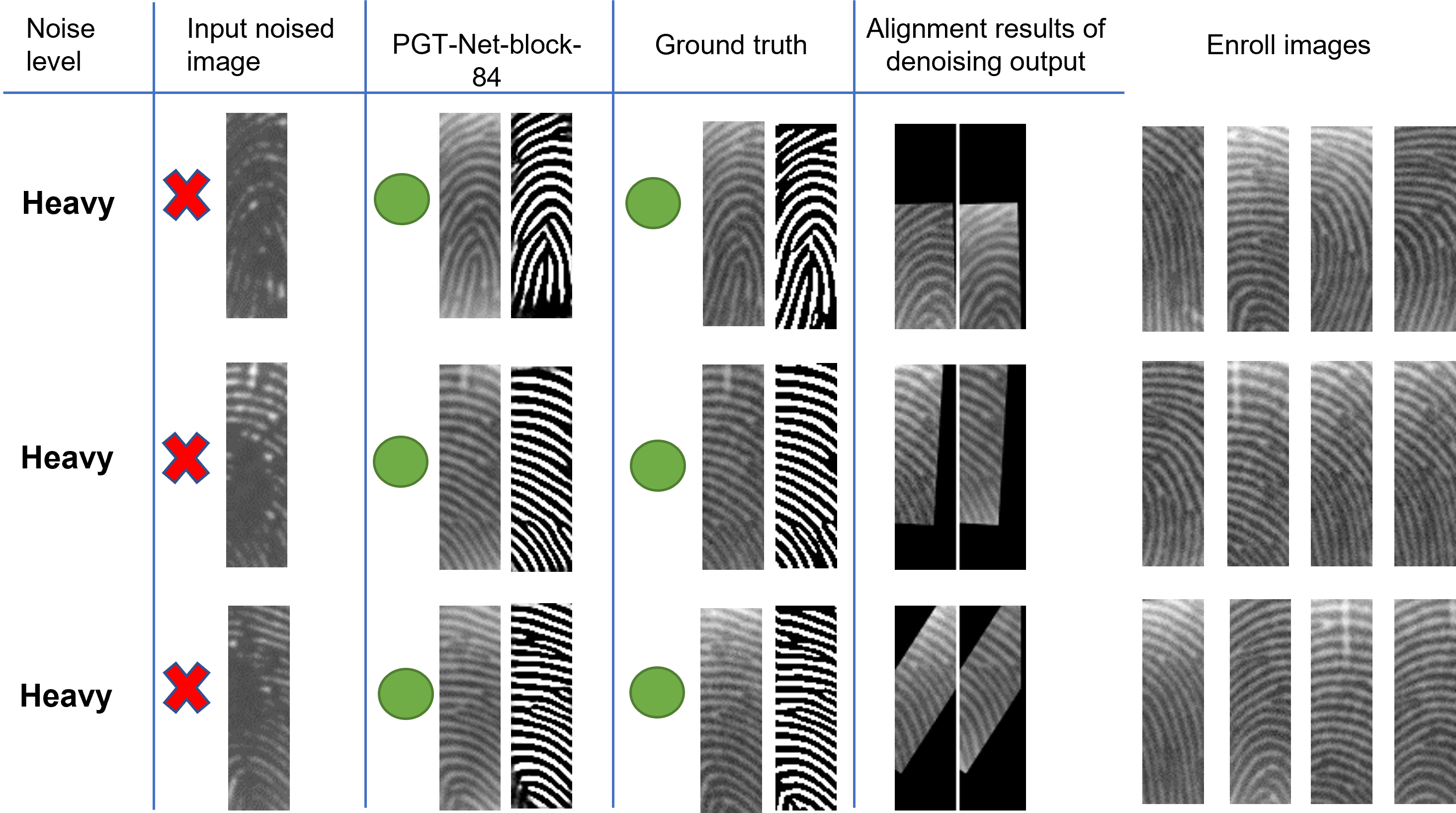}
\caption{PGT-Net-block-84 denoised and recognition
results on the FT-lightnoised dataset (heavy noise).}
\label{fig:ft_light_summarize_img_h}
\end{figure}

\begin{figure}[!t]
\centering
\includegraphics[width=\linewidth]{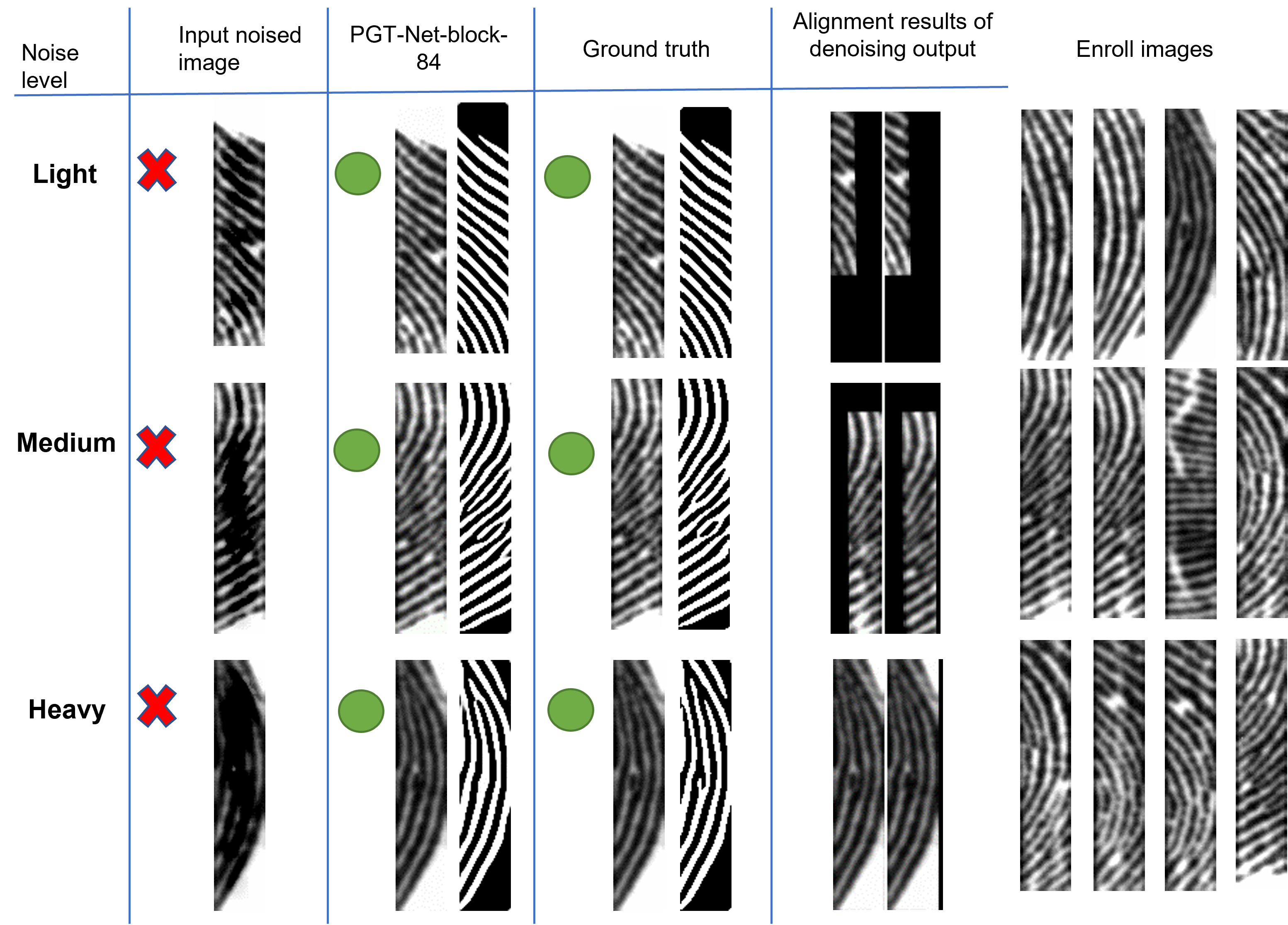}
\caption{PGT-Net-block-84 denoised images and recognition results on the FW9395 dataset.}
\label{fig:FW9395_summarize_img}
\end{figure}

\begin{figure}[!t]
\centering
\includegraphics[width=\linewidth]{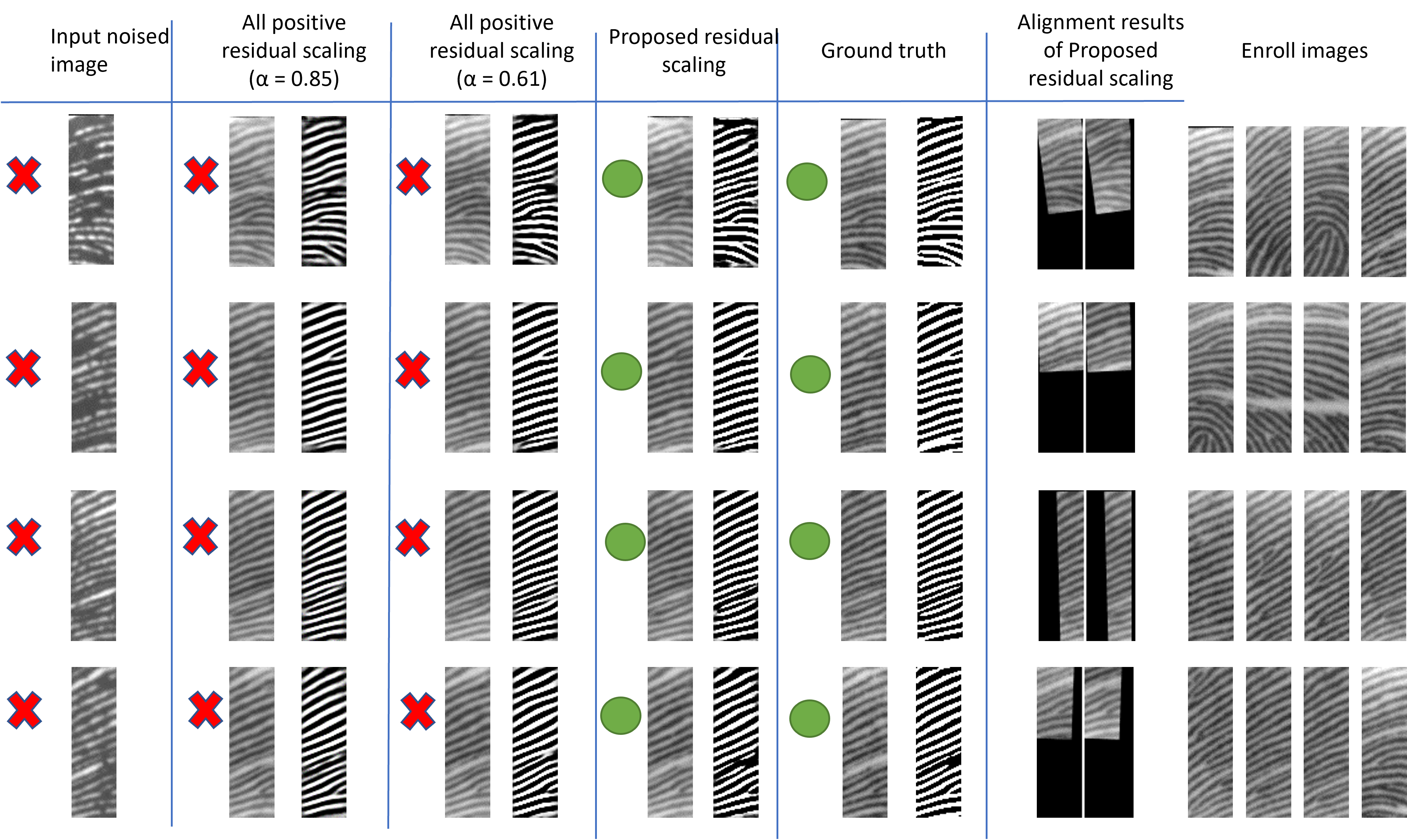}
\caption{The denoised images and recognition results with different residual scaling settings.}
\label{fig:residual_scaling_output_img}
\end{figure}

\renewcommand{\arraystretch}{1.5}
\begin{table}[h!]
    \centering
    \caption{Detailed information of recognition datasets.}
    \label{tab:recognition_dataset}
    \begin{adjustbox}{width=\columnwidth, center}
    
    \begin{tabular}{|c|c|c|c|c|c|}     
        \hline
        Dataset & \makecell{Person\\ ID} & \makecell{\# of\\ enrolled\\ images} & \makecell{\# of\\ identified\\ images} & \makecell{\# of\\ successful\\ identification} & FRR(\%)\\
        \hline
        \multirow{5}{*}{FT-lightnoised} & 0001 & 96 & 1145 & 797 & 30.4 \\
        
        & 0002 & 96 & 903 & 714 & 20.9 \\
        & 0003 & 96 & 597 & 478 & 19.9 \\
        & 0004 & 96 & 1781 & 1651 & 7.29 \\
        \cline{2-6}
        & Total & 384 & 4426 & 3640 & 17.8 \\
        \hline
        
        \multirow{7}{*}{\makecell{FW9395}} 
        & 0001 & 24 & 71 & 64 & 9.85 \\
        
        & 0057 & 24 & 65 & 58 & 10.76 \\
        & 0060 & 24 & 78 & 68 & 12.82 \\
        & 0063 & 24 & 84 & 77 & 8.33 \\
        & 0064 & 24 & 80 & 76 & 5.00 \\
        & 0103 & 24 & 77 & 69 & 10.38 \\
        \cline{2-6}
        & Total & 144 & 455 & 412 & 9.45 \\
        \hline
    \end{tabular}
    \end{adjustbox}
\end{table}

\section{Experimental Results}
\label{ch:results}

\subsection{Performance Summarization}
\label{sec:Performance_summarization}
Table~\ref{tab:denoising_performance_summarization_ft_lightnoised} summarizes the denoising performance of different models. Table~\ref{tab:84_recognition_performance_summarization} summarizes the recognition performance of different models. Note that because the PGT-Net-Edge is optimized for the FW9395 dataset, it may perform poorly on the FT-lightnoised dataset, which has more complicated noise. Also, since the FW9395 dataset has simpler noise synthesized artificially, the differences between single-task and multi-task are not very much. For the more complicated FT-lightnoised dataset, we can see the value of the proposed multi-task model.

To summarize, Fig.~\ref{fig:ft_light_summarize_img_lm} \& Fig.~\ref{fig:ft_light_summarize_img_h} show the results for the FT-lightnoised dataset. Fig.~\ref{fig:FW9395_summarize_img} shows the results for the FW9395 dataset. The three figures mentioned above are all based on PGT-Net-block-84, the model with the best performance. As you can see from those figures, the proposed algorithm has an outstanding result, with different noise levels or sensors.

As mentioned in section ~\ref{ch:PGT_Net_on_edge_devices}, the PGT-Net-Edge reduces the complexity be three ways described below. 

\begin{itemize}
\item Residual blocks quantity reduction.
\item Residual blocks output channel reduction.
\item Dynamic fixed point quantization.
\end{itemize}

Table~\ref{tab:PGT-Net-Edge_FW9395} shows the experimental results after quantization. There are few performance downgrades, but the model size has been quartered after quantization from float32 to the 8-bit dynamic-fixed point.
As you can see from the tables, there are only a few performance downgrades compared to the  PGT-Net-block-84, which have similar dataflow but a higher quantity of parameters.

\renewcommand{\arraystretch}{1.8}
\begin{table}[h!]
    \centering
    \caption{Denoising performance summarization}
    \label{tab:denoising_performance_summarization_ft_lightnoised}
    \begin{adjustbox}{width=\linewidth, center}
    \begin{tabular}{|c|c|m{4em}|m{4em}|m{4em}|}     
        \hline
        Dataset & Model & MSE & SSIM & PSNR\\
        \hline
        \multirow{5}{*}{FT-lightnoised} 
        & Single-task PGT-Net-block-84 & 0.0048 & 0.9172 & 24.1606 \\
        & PGT-Net-Edge & 0.0103 & 0.8605 & 20.3622 \\
        & Multi-task PGT-Net-block-84 & \textbf{0.0044} & \textbf{0.9270} & \textbf{24.7407} \\
        &  No enhance & 0.0204 & 0.6997 & 17.6251 \\
        \cline{2-5}
        & Improvement(\%) & 78.13 & 32.49 & 40.37 \\

        \hline
        \multirow{5}{*}{FW9395} 
        & Single-task PGT-Net-block-84 & 0.0004 & \textbf{0.9894} & \textbf{35.0790} \\
        & PGT-Net-Edge & 0.0008 & 0.9793 & 31.8389 \\
        & Multi-task PGT-Net-block-84 & \textbf{0.0004} & 0.9891 & 34.8153 \\
        &  No enhance & 0.0236 & 0.7780 & 16.4354 \\
        \cline{2-5}
        & Improvement(\%) & 98.16 & 27.13 & 111.83 \\
        \hline

        \multicolumn{5}{l}{\footnotesize $\text{Improvement(\%)} = \frac{\text{(result)}-\text{(No enhance)}}{\text{(No enhance)}} \times 100\%$}\\
    \end{tabular}
    \end{adjustbox}
\end{table}

\renewcommand{\arraystretch}{2}
\begin{table}[h]
    \centering
    \begin{normalsize}
    \caption{Recognition performance summarization for the FT-lightnoised and FW9395 dataset.}
    \label{tab:84_recognition_performance_summarization}
    \begin{adjustbox}{width=\linewidth, center}
    \begin{tabular}{|c|c|c|c|c|c|}     
        \hline
        Dataset & \makecell{\# of\\ enrolled\\ images} & \makecell{\# of\\ identified\\ images}& model & \makecell{\# of\\ successful\\ recognition} & FRR(\%)\\
        \hline
        \multirow{4}{*}{\makecell{FT-lightnoised}} & \multirow{4}{*}{384} & \multirow{4}{*}{4426} & No enhance & 3640 & 17.75 \\
        \cline{4-6}
        & &  & Single-tasks PGT-Net-block-84 & 4199 & 5.12 \\
        \cline{4-6}
        & &  & PGT-Net-Edge & 3444 & 22.18 \\
        \cline{4-6}
        & &  & Multi-task PGT-Net-block-84 & 4228 & 4.47 \\
        \hline
        
        \multirow{4}{*}{\makecell{FW9395}} & 
        \multirow{4}{*}{144} & \multirow{4}{*}{455} & No enhance & 412 & 9.45 \\
        \cline{4-6}
        &  & & Single-tasks PGT-Net-block-84 & 443 & 2.64 \\
        \cline{4-6}
        &  & & PGT-Net-Edge & 445 & 2.19 \\
        \cline{4-6}
        &  & & Multi-task PGT-Net-block-84 & 450 & 1.09 \\
        \hline
    \end{tabular}
    \end{adjustbox}
    \end{normalsize}
\end{table}

\renewcommand{\arraystretch}{1.8}
\begin{table}[t!]
    \centering
    \caption{PGT-Net-Edge experimental results on the FW9395 dataset.}
    \label{tab:PGT-Net-Edge_FW9395}
    \begin{adjustbox}{width=\linewidth, center}
    \begin{tabular}{|c|c|c|c|c|}    
        \hline
        Model & PGT-Net-block-84 & \multicolumn{3}{c|}{PGT-Net-Edge} \\
        \hline
        \makecell{Quantization\\precision} &  N/A  &  N/A  & \makecell{8 bits dynamic\\ fixed point} & \makecell{8 bits dynamic\\ fixed point} \\ 
        \hline 
        \makecell{Quantized\\weight/activations} & N/A & N/A & Weight only & \makecell{Weight \& input activations \\ \& output activations} \\
        \hline
        SSIM & 0.9891 & 0.9875 & 0.9873 & 0.9793 \\
        \hline
        MSE  & 0.0004 & 0.0005 & 0.0006 & 0.0008 \\
        \hline
        PSNR & 34.8153 & 33.0903 & 32.6867 & 31.8389 \\
        \hline
        \# of parameters & 6,636,994 & \multicolumn{3}{c|}{687,042} \\
        \hline
        Size & 26.547MB & \multicolumn{3}{c|}{0.687MB} \\
        \hline
    \end{tabular}
    \end{adjustbox}
\end{table}

\subsection{Performance Comparison}
\label{sec:Performance_comparison}
Table~\ref{tab:Denoising_results_of_related_works} summarizes the experiment results of the fingerprint denoising-related works mentioned above. Table~\ref{tab:Recognition_results_of_related_works} summarizes the experiment results of fingerprint recognition compared with related works.
According to these two table, we observe that PGT-Net performs well on denoising and recognition tasks. From Table~\ref{tab:Recognition_results_of_related_works}, PGT-Net yields substantial improvement in fingerprint recognition tasks, with a minor increase in the FAR. The FAR slightly elevate from 0.02\% to 0.07\%. One potential explanation is that during the restoration process of our proposed model, features belonging to fingerprints from other classes are inadvertently restored, as PGT-Net did not incorporate fingerprint categories in its training process. We consider this to be an acceptable trade-off when weighed against the enhancement in FRR.

\begin{table}[h!]
    \centering
    \caption{Denoising results of related works}
    \label{tab:Denoising_results_of_related_works}
    \begin{adjustbox}{width=0.5\textwidth, center}
    \begin{tabular}{|c|c|c|m{5em}|c|c|c|}
    \hline
    Dataset                         & Model            & Note                                               & Application                  & MSE      & SSIM     & PSNR      \\ \hline
    \multirow{7}{*}{\makecell{\\\\\\\\\\\\\\\\\\FT-lightnoised}} & \makecell{DenseUNet\\~\cite{Latent_Fingerprint_Enhancement_Based_on_DenseUNet}(2019)}        & \makecell{Based on\\ U-Net}                                     & Latent fingerprint denoising & 0.0125 & 0.8592 & 19.7396 \\ \cline{2-7} 
                                & \makecell{PFE-Net\\~\cite{FENET}(2019)}          & \makecell{Based on\\ Residual\\ stucture}                         & Fingerprint pore matching    & 0.0122  & 0.7567 & 19.5793 \\ \cline{2-7} 
                                & \makecell{FingerNet\\~\cite{Deep_convolutional_neural_network_for_latent_fingerprint_enhancement}(2018)}        & \makecell{Based on\\ Multi-task\\ learning}                      & Latent fingerprint denoising & 0.0130    & 0.8687   & 20.0429   \\ \cline{2-7} 
                                & \makecell{FPD-M-Net\\~\cite{FPDMNET}(2019)}          & \makecell{Based on\\ M-Net~\cite{MNet}}                                     & Latent fingerprint denoising & 0.0097   & 0.7990    & 20.8516   \\ \cline{2-7} 
                                & \makecell{Residual M-net\\~\cite{cunha2022residual}(2022)}   & \makecell{Based on\\ M-Net~\cite{MNet}}                                     & Latent fingerprint denoising & 0.0210    & 0.6452   & 17.0903   \\ \cline{2-7} 
                                & PGT-Net-Block-84 & \makecell{Based on Residual\\ structure \& Multi\\-tasks learning} & Wet fingerprint denoising    & \textbf{0.0044}   & \textbf{0.9270}   & \textbf{24.7400}    \\ \cline{2-7} 
                                & No enhance       & N/A                                                & \makecell{\,\,N/A}                          & 0.0204   & 0.6997   & 17.6251   \\ \hline
\end{tabular}
    \end{adjustbox}
   
\end{table}

\renewcommand{\arraystretch}{1.8}
\begin{table}[h!]
    \centering
    \caption{Recognition results of related works}
    \label{tab:Recognition_results_of_related_works}
    \begin{adjustbox}{width=0.5\textwidth, center}
    \begin{tabular}{|c|c|c|c|c|c|c|}
\hline
Dataset                         & \makecell{\# of\\ enrolled\\ images} & \makecell{\# of\\ identified\\ images} & Model            & \makecell{\# of\\ successful\\ recognition} & FRR(\%)           & FAR(\%)  \\ \hline
\multirow{7}{*}{FT-lightnoised} & \multirow{7}{*}{384}  & \multirow{7}{*}{4428}   & \makecell{DenseUNet~\cite{Latent_Fingerprint_Enhancement_Based_on_DenseUNet}(2019)}        & 3975                         & 10.26         & 0.02 \\ \cline{4-7} 
                                &                       &                         & \makecell{PFE-Net~\cite{FENET}(2019)}          & 3565                         & 19.45         & 0.02 \\ \cline{4-7} 
                                &                       &                         & \makecell{FingerNet~\cite{Deep_convolutional_neural_network_for_latent_fingerprint_enhancement}(2018)}        & 3785                         & 14.48         & 0.05 \\ \cline{4-7} 
                                &                       &                         & \makecell{FPD-M-Net~\cite{FPDMNET}(2019)}          & 2793                         & 36.9          & 0.02 \\ \cline{4-7} 
                                &                       &                         & \makecell{Residual M-net~\cite{cunha2022residual}(2022)}   & 3071                         & 30.62         & 0.02 \\ \cline{4-7} 
                                &                       &                         & PGT-Net-Block-84 & \textbf{4228}                & \textbf{4.47} & 0.07 \\ \cline{4-7} 
                                &                       &                         & No enhance       & 3640                         & 17.8          & 0.02 \\ \hline
\end{tabular}
    \end{adjustbox}
\end{table}

\subsection{Ablation Studies}
\label{sec:Ablation_Studies}
Here we present ablation experiments to analyze the contribution
of each component of our model. Evaluation is performed on the FW9395 and FT-lightnoised datasets.

\subsubsection{Single-task versus Multi-task Model}

\begin{figure}[!t]
\centering
\subfloat[]{\includegraphics[height=4.2cm ]{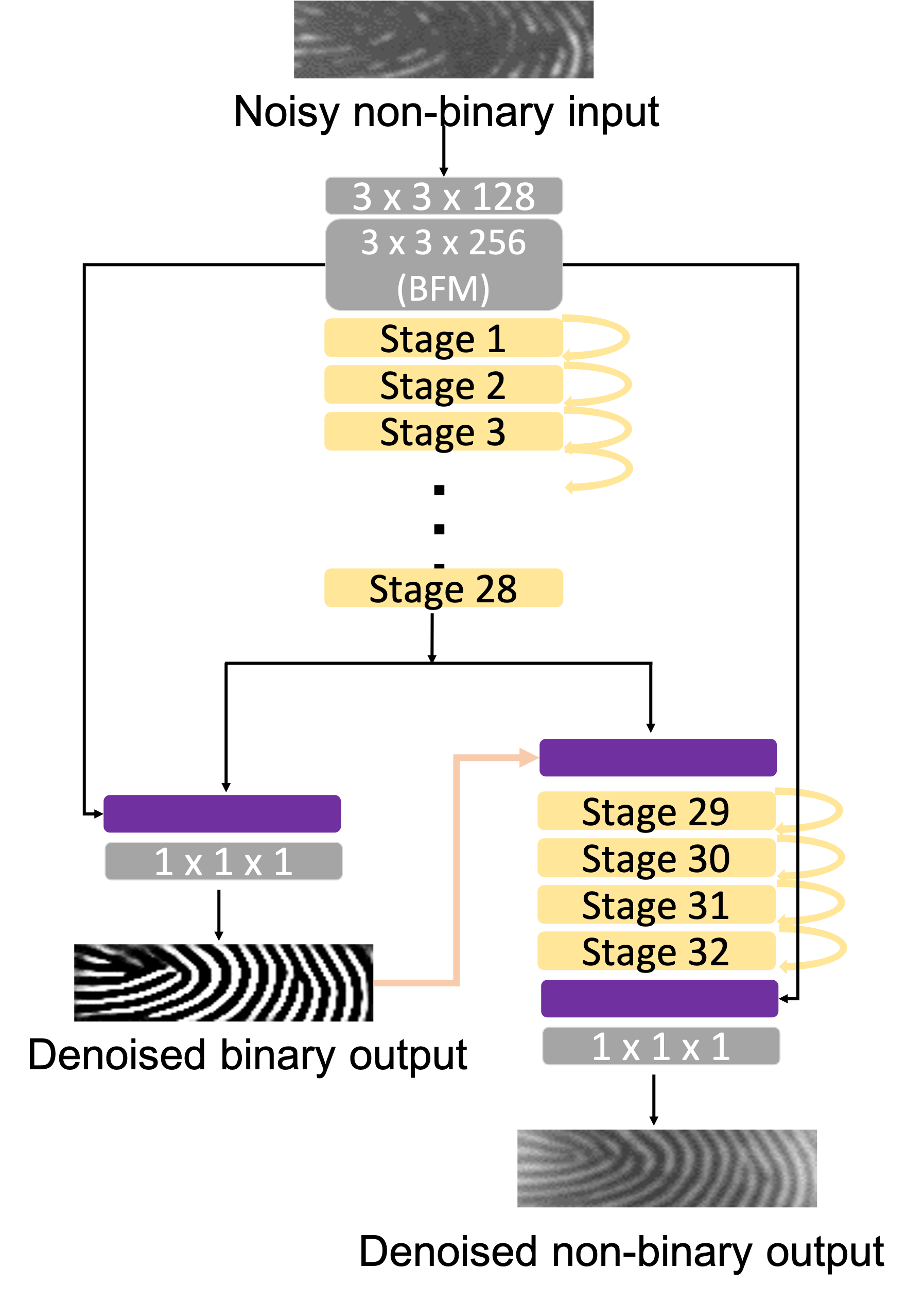}%
\label{fig:PGT_Net_Edge}}
\hfil
\subfloat[]{\includegraphics[height=4.2cm]{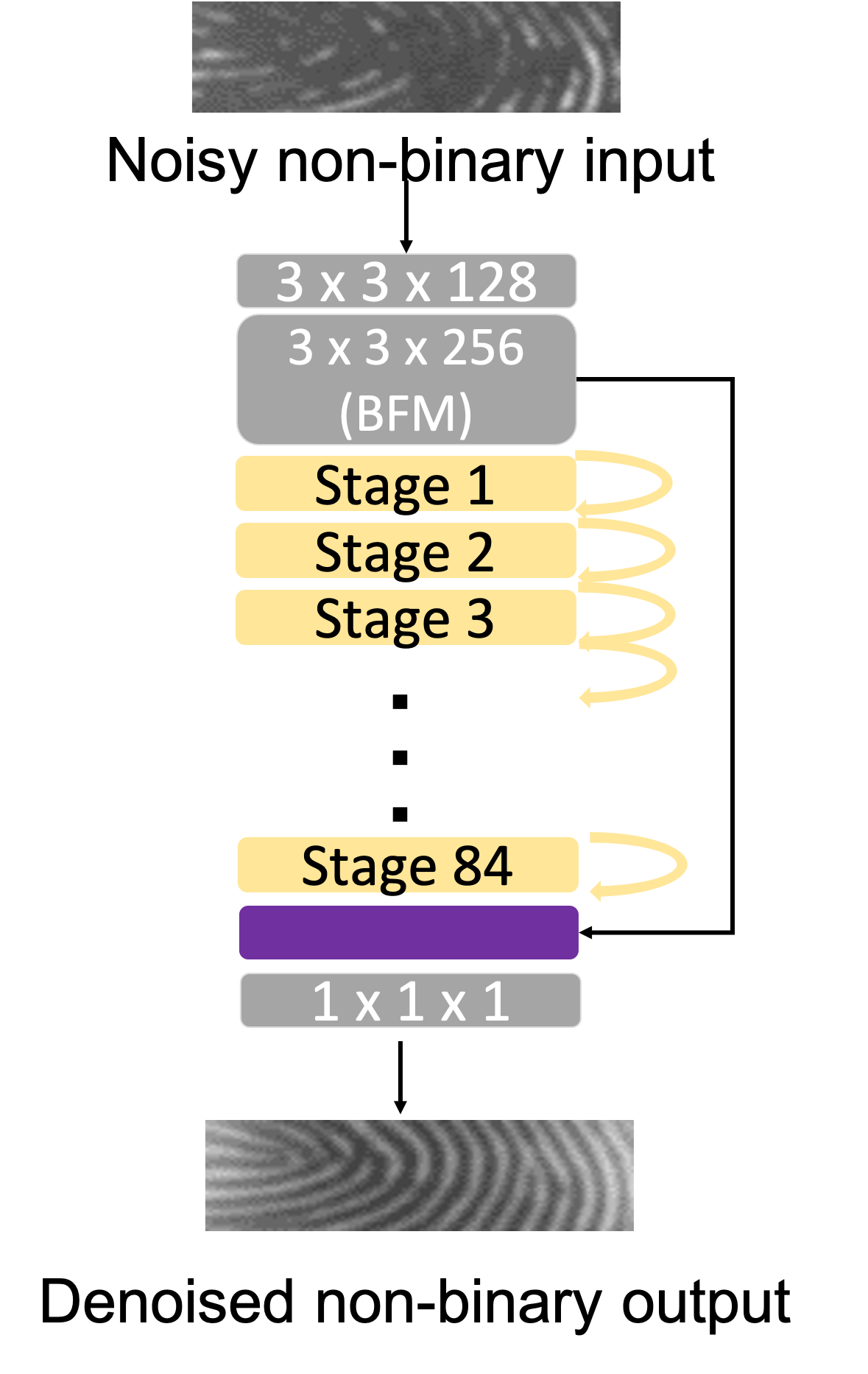}%
\label{fig:PGT_single_task}}
\hfil
\subfloat[]{\includegraphics[height=5.2cm]{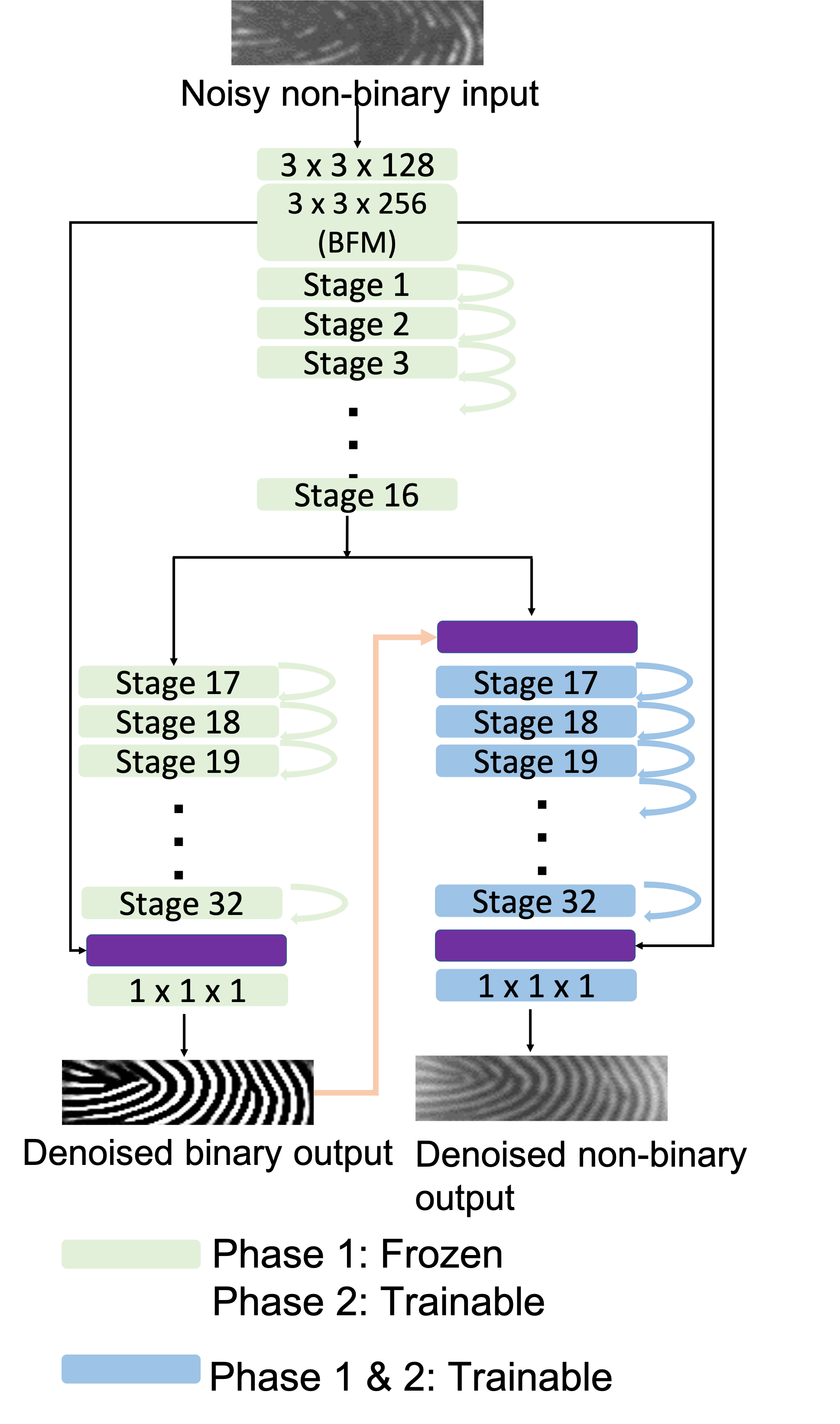}%
\label{fig:Two Phase PGT-Net}}
\caption{PGT-Net variants. (a) The architecture of PGT-Net-Edge (b) Single-task
version of PGT-Net-block-84 (c) Two-phase training process of PGT-Net}
\label{fig:PGT_Net_Edge_and_PGT_single_task}
\end{figure}

Our model yields better performance as Multi-task model(Fig.~\ref{fig:PGT-Net-block-84_dataflow}). Fig.~\ref{fig:PGT_Net_Edge_and_PGT_single_task} shows an architecture of the PGT-Net-single-task; it is similar to the original PGT-Net model but in the single-task version, and the number of the residual blocks are the same, so they have a similar quantity of parameters. Also, both single-task and multi-task have the same residual scaling setting with \(\alpha\) = 24, as described in Eq.~\ref{eq:eplison}, and thus a fair comparison. We demonstrate the evaluation results of single-task and Multi-task model training with FT-lightnoised dataset. 

Table ~\ref{tab:denoising_performance_summarization_ft_lightnoised} shows the recognition performance of the single-task and Multi-task PGT-Net-block-84. As you can see from the table, multi-task outperforms single-task and have a lower FRR in fingerprint recognition.

According to Fig.~\ref{fig:multisingle_FPDM_result}, we can also observe that both single-task and multi-task perform well on denoising the noisy fingerprint from the first column. Still, in some details, multi-task has a more accurate fingerprint contour (marked in red boxes), and such information makes the denoised fingerprint of multi-task recognized successfully. In conclusion, the binary branch did help in fingerprint restoration.

\subsubsection{Binary Progressive Guided Task}
Here we use some techniques to improve the performance of our Multi-task model. 
\begin{itemize}
\item \textit{Progressive Guided Concatenation Path:} The model with the ``binary guided concatenate path" is shown in Fig.~\ref{fig:PGT-Net-block-84_dataflow}, and the model without the “Progressive guided concatenation path” is modified from the original PGT-Net with the “Progressive guided concatenation path” begin deleted. Experiments shows that the concatenate path can improve the performance.
\item \textit{Separate the Training Process:}
\begin{itemize}
\item Phase 1: Train the binary-related parameters in our model first. 
\item Phase 2: Do the regular training.
\end{itemize}
Such a technique can improve the performance because the binary path in the model is pre-trained and can produce a more precise binary fingerprint contour to guide the model toward better performance more accurately.

Fig.~\ref{fig:Two Phase PGT-Net} shows the partial model that can be trained during the 2-Phase training process. Yellow blocks are trainable both in Phases 1 and 2. Gray blocks are only trainable in Phase 2.
\end{itemize}

Table~\ref{tab:bpgt} summarizes the denoising performance with different settings. As you can observe from the table, the “progressive guided concatenation path” and “progressive 2-stage training” can make the model performs better.

\begin{table}[h!]
    \centering
    \caption{Ablation study on binary progressive guided multi-task and 2-phase training on FT-lightnoised.}
    \label{tab:bpgt}
    \begin{adjustbox}{width=\linewidth, center}
    \begin{tabular}{|c|ccc|c|c|c|}
    \hline
    Model &   Multi-task & Progressive guided & 2 Phase training & MSE & SSIM & PSNR\\
    \hline
    \multirow{4}{*}{\makecell{PGT-Net-\\block-84}} & V  &   &   &   0.0058 & 0.9021 & 23.0740 \\
        &   V   &   V   &      &    0.0056 & 0.9090 & 23.3149 \\
        &   V   &   V   &   V   &    \textbf{0.0053} & \textbf{0.9121} & \textbf{23.5583} \\
    \cline{2-7}
       & Improvement (\%) &     &   & 9.089   &   1.112  &   2.099 \\
    \hline
    \multicolumn{4}{l}{\footnotesize $\text{Improvement(\%)} = \frac{\text{(result)}-\text{(No enhance)}}{\text{(No enhance)}} \times 100\%$}\\
    \end{tabular}
    \end{adjustbox}
\end{table}

\subsubsection{Residual scaling factor \(\varepsilon\)}
\label{sec:Residual_scaling_factor}

As described previously in section~\ref{Residual_blocks}
\begin{itemize}
\item  Each residual block has a constant \(\varepsilon\) for residual scaling, as shown in Fig.~\ref{fig:residualBlock}
\item The scaling factor \(\varepsilon\) can be either positive or negative, as shown in Fig.~\ref{fig:PGT-Net-block-84_scale}.
\item In our work, we set \(\alpha\) to 24 in Eq.~\ref{eq:eplison}
\end{itemize}

If we set the \(\varepsilon\) as the FENet~\cite{FENET}, such that all the residual blocks have positive \(\varepsilon\), namely \(\varepsilon = 0.01 \times \alpha - current\;stage)\), with \(\alpha = 61\) or \(85\), the training loss will significantly increase, and thus a poor recognition performance.

There are two different settings with positive \(\varepsilon\), as described below:
\begin{itemize}
\item  \(\alpha = 61\) because the maximum stage of PGT-Net-block-84 is 60. And the model architecture of this setting is exactly the same as PGT-Net-block-84 (Fig.~\ref{fig:PGT-Net-block-84_dataflow}).

\item \(\alpha = 85\) because there are 84 residual blocks in PGT-Net-block-84. Although the model architecture is similar to PGT-Net-block-84, it still exists some differences. The stage of the non-binary branch starts at 61 instead of 25, ends at stage 84 instead of 48, as shown in Fig.~\ref{fig:Two Phase PGT-Net}. The non-binary branch starts at 61 because it has the input comes from the binary output, which ends at stage 60.
\end{itemize}

\begin{figure}[ht]
    \centering
    \includegraphics[width=0.75\linewidth]{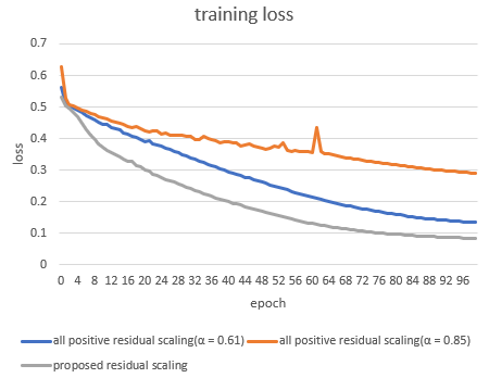}
    \caption{Training loss of different residual scaling setting.}
    \label{fig:residual_scaling_training_loss}
\end{figure}

The training loss of the residual scaling setting mentioned above is shown in Fig.~\ref{fig:residual_scaling_training_loss}. The proposed residual scaling can further reduce the training loss and thus achieve better denoising and recognition performance.

Table~\ref{tab:denoising_performance_residual_scaling} summarizes the denoising performance for the FT-lightnoised dataset, and Table~\ref{tab:residual_scaling_reg} summarizes the recognition performance for the FT-lightnoised dataset. The proposed residual scaling can indeed perform better in our experiments.

Fig.~\ref{fig:residual_scaling_output_img} shows the denoised output of different residual scaling settings. Those images denoised with all positive residual scaling settings can’t be recognized successfully, but the output fingerprint of proposed residual scaling can. Although all of them can produce denoised clear fingerprint output, on some details (marked in red boxes), proposed residual scaling makes the denoised fingerprint texture more precise and thus have lower FRR.

\begin{table}[h!]
    \centering
    \caption{Ablation study on different residual scaling settings for the FT-lightnoised dataset.}
    \label{tab:denoising_performance_residual_scaling}
    \begin{adjustbox}{width=\linewidth, center}
    \begin{tabular}{|c|c|c|c|c|}     
        \hline
        Model & Residual scaling settings & MSE & SSIM & PSNR\\
        \hline
        \multirow{3}{*}{\makecell{PGT-Net-\\block-84}} & \makecell{all positive scaling (\(\alpha\) = 0.85)} & 0.0061 & 0.8957 & 22.8687 \\
        \cline{2-5}
        & \makecell{all positive scaling(\(\alpha\) = 0.61)} & 0.0051 & 0.9111 & 23.7630\\
        \cline{2-5}
        & \makecell{proposed residual scaling} & {\textbf{0.0044}} & {\textbf{0.9270}} & {\textbf{24.7407}} \\
        \hline
    \end{tabular}
    \end{adjustbox}
\end{table}

\renewcommand{\arraystretch}{1.8}
\begin{table}[t!]
    \centering
    \caption{Recognition performance of different residual scaling settings using FT-lightnoised dataset.}
    \label{tab:residual_scaling_reg}
    \begin{adjustbox}{width=\linewidth, center}
\begin{tabular}{|c|c|c|c|c|c|}     
        \hline
        model & \makecell{\# of\\ enrolled\\ images}  & \makecell{\# of\\ identified\\ images}  & model & \makecell{\# of\\ successful\\ recognition} & \makecell{FRR\\(\%)}\\   
        \hline
        \multirow{4}{*}{\makecell{PGT-Net-\\block-84}} & \multirow{4}{*}{384} & \multirow{4}{*}{4426} & No enhance & 3640 & 17.75 \\
        \cline{4-6}
        &  &  & All positive scaling(\(\alpha\) = 0.85) & 3910 &  11.65  \\
        \cline{4-6}
        &  &  & All positive scaling(\(\alpha\) = 0.61) & 4100 & 7.36  \\
        \cline{4-6}
        &  &  & {\textbf{Proposed residual scaling}} & {\textbf{4228}} & {\textbf{4.47}}\\
        \hline
\end{tabular}
\end{adjustbox}
\end{table}

\section{Conclusion}
\label{chap:Conclusion}
In this work, we presented PGT-Net for wet fingerprint denoising. The proposed methodologies have proven to be effective in multiple types of sensors. And can recover the noisy fingerprints covered with real and synthetic noise. PGT-Net uses residual blocks as the fundamental structures. And considering that the fingerprints get by the sensor would be small and thin, multi-task architecture is added in PGT-Net to use the results of supported tasks to guide the main task to reach better performance. Also, the proposed residual scaling did a great job at reducing the training loss.

The proposed data flow is simple and scalable. One can easily modify the model according to the proposed data flow and will get a result with an excellent denoising performance.

The convenience of fingerprint recognition in our daily usage has also been greatly improved. The proposed methodologies have proven effective in both optical and capacitive sensors. The FRR has been reduced from 17.75 \% to 4.47 \% for the FT-lightnoised dataset. FRR has been reduced from 9.45 \% to 1.09 \% for the FW9395 dataset.

\bibliographystyle{IEEEtran}
\bibliography{Reference}

\end{document}